\documentclass[10pt,twocolumn,letterpaper]{article}

\usepackage{iccv}
\usepackage{times}
\usepackage{epsfig}
\usepackage{graphicx}
\usepackage{amsmath}
\usepackage{amssymb}

\usepackage[breaklinks=true,bookmarks=false]{hyperref}

\iccvfinalcopy 

\def\iccvPaperID{****} 

\ificcvfinal
\newcommand*{\affaddr}[1]{#1}
\newcommand*{\affmark}[1][*]{\textsuperscript{#1}}
\newcommand*{\email}[1]{\tt\small{#1}}
\begin{document}

\title{The Third Place Solution for CVPR2022 AVA Accessibility Vision and Autonomy Challenge}

\author{Bo Yan\affmark[1], Leilei Cao\affmark[1,2], Zhuang Li\affmark[1,3], Hongbin Wang\affmark[1] \\
\small
\affaddr{\affmark[1]Ant Group} \quad 
\affaddr{\affmark[2]Northwestern Polytechnical University} \quad
\affaddr{\affmark[3]Tongji University} \quad
\\
\email{lengyu.yb@antgroup.com, hongbin.whb@antgroup.com} \\
}

\maketitle

\begin{abstract}

The goal of AVA challenge is to provide vision-based benchmarks and methods relevant to accessibility. In this paper, we introduce the technical details of our submission to the CVPR2022 AVA Challenge. Firstly, we conducted some experiments to help employ proper model and data augmentation strategy for this task. Secondly, an effective training strategy was applied to improve the performance. Thirdly, we integrated the results from two different segmentation frameworks to improve the performance further. Experimental results demonstrate that our approach can achieve a competitive result on the AVA test set. Finally, our approach achieves 63.008\%AP@0.50:0.95 on the test set of CVPR2022 AVA Challenge.

\end{abstract}

\section{Introduction}

Instance segmentation applies widely in image editing, image composition, autonomous driving, etc. Instance segmentation is a fundamental problem in computer vision. Deep learning-based methods have achieved promising results for image instance segmentation over the past few years, such as Mask R-CNN~\cite{he2017mask}, PANet~\cite{liu2018path}, TensorMask~\cite{chen2019tensormask},  CenterMask~\cite{wang2020centermask}, SOLO series~\cite{wang2020solo, wang2020solov2} . Specifically, AVA challenge involves a synthetic instance segmentation benchmark incorporating use-cases of autonomous systems interacting with pedestrians with disabilities~\cite{9710390}.

In addition, transformers have made enormous strides in NLP\cite{devlin2018bert, radford2019language}. There are quite a bit of works applying transformers to computer vision\cite{ramachandran2019stand, Zhao_2020_CVPR, carion2020end} because transformers can capture the non-local and relational nature of images. Especially, Swin Transformer\cite{liu2021swin} has been widely used for many computer vision tasks and achieves successful results, such as detection and segmentation task on COCO.

\hfill

In order to address the AVA instance segmentation task, firstly we conducted some experiments to test whether previous studies are effective for this task, which can help us employ proper model and data augmentation strategy. Secondly, an effective training strategy is applied to improve the model performance. Finally, we integrated the results from two different segmentation frameworks to improve the performance further.

\hfill

\section{Approach}

Our approach mainly includes three parts: segmentation model and data augmentation, training strategy, and model integration. We introduce the segmentation model and data augmentation strategy in Sec.2.1. The training strategy is introduced in Sec.2.2. Moreover, the details of model integration is introduced in Sec.2.3.

\subsection{Model and Data}

Our segmentation model is Hybrid Task Cascade(HTC)~\cite{chen2019hybrid} based detector on the CBSwin-Large backbone with CBFPN~\cite{Liang2021CBNetV2AC}.

Copy-Paste~\cite{9578639}, which copies object from one image to another, is particularly useful for instance segmentation. There are only eight categories in AVA challenge, and the number and area of instance per image are small in many cases, so Copy-Paste is a very effective data augmentation for AVA challenge.

\begin{figure*}[ht]
	\centering
	\includegraphics[scale=0.425]{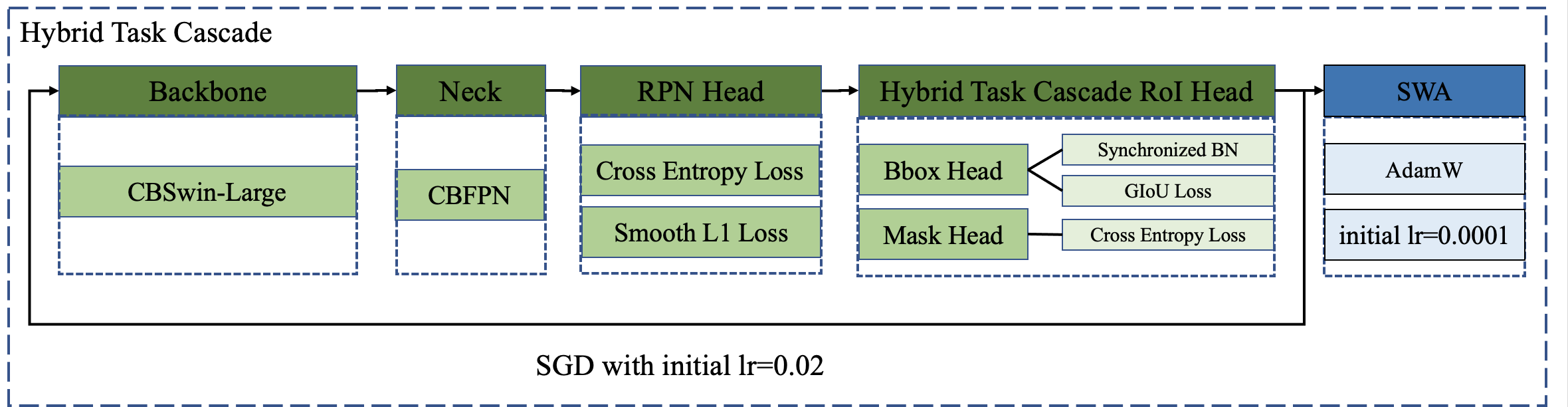}
	\caption{Our model architecture and training pipeline for CVPR2022 AVA Challenge.}
	\label{pipeline}
\end{figure*}

\begin{table*}[ht]
\begin{center}
\begin{tabular}{lcc}
\hline
Methods & AP@0.50:0.95 & Boost \\
\hline
HTC-CBSwin-Large + Soft-NMS + Flip & 58.293\% & - \\
HTC-CBSwin-Large + SWA + Soft-NMS + Flip & 59.852\% & 1.559\% \\
HTC-CBSwin-Large + Copy-Paste + Soft-NMS + Flip & 60.569\% & 0.717\% \\
HTC-CBSwin-Large + Copy-Paste + SWA + Soft-NMS + Flip & 62.603\% & 2.034\% \\
HTC-CBSwin-Large + Copy-Paste + SWA + Soft-NMS + Flip\\+ Mask2Former & 63.008\% & 0.405\% \\
\hline
\end{tabular}
\end{center}
\caption{Experimental results on test set of CVPR2022 AVA Challenge.}
\end{table*}

\subsection{Training Strategy}

Firstly, we train the model with Copy-Paste data augmentation. Stochastic Gradient Descent(SGD)is applied to this stage. After the model converged, we use SWA~\cite{zhang2020swa} training strategy to finetune the model, which can make the model better and more robust. Adam with decoupled weight decay(AdamW)~\cite{loshchilov2018decoupled} is applied during the SWA training stage.

Our model architecture and training pipeline are shown in Figure 1.

\subsection{Model Integration}
 Mask2Former~\cite{cheng2021mask2former} is a new architecture capable of addressing any image segmentation task, include  panoptic, instance and semantic. Mask2Former is very different from the previous instance segmentation framework. In order to combine both of advantages of Mask2Former and HTC, we directly train the Mask2Former in AVA dataset without in-depth research, and integrated the results from HTC and Mask2Former as the final results.

\hfill

\section{Experiments}

We train and evaluate our approach on the CVPR2022 AVA Challenge dataset.  

\subsection{Training Details}
We use the AVA2022 train and validation dataset to train and evaluate the model. In the first stage of training, the pre-trained CBSwin-Large model by COCO is applied, we train the model with SGD and initial lr=0.02. Then the SWA training strategy is applied to finetune the model, and the optimizer is AdamW with initial lr=0.0001. 

The input images are randomly scaled from 720 to 1620 on the short side and up to 1920 on the long side. Then randomly cropped and padded to [1920, 1080]. Finally, random flip and  Copy-Paste data augmentation methods are applied, the augmented images are input to the model for training.

\subsection{Experimental Results}

As shown in Table 1, our approach finally achieves 63.008\%AP@0.50:0.95 on the CVPR2022 AVA instance segmentation challenge test set.


\subsection{Ablation Study}

This section elaborates on how we achieve the final result by ablation study to explain our approach. The baseline is HTC-CBSwin-Large, and Copy-Paste is very effective data augmentation method for this task. In order to improve the recall of the model, Soft-NMS~\cite{bodla2017soft} is used on the test stage for all experiments. And test time augmentation(TTA) is also applied, we experiment if add flip horizontal for every image, it can achieve a better score, but for multi scale test with scale factors [0.5,1.0,2.0] or [0.8,1.0,1.2], it can't bring any improvement. Then we use SWA training strategy to finetune the model, it can bring an improvement of 2.034\%. And we also train the Mask2Former in AVA train dataset without in-depth research, we find the segment results of Mask2Former is worse than HTC-CBSwin-Large except category "cane", so we integrate the results from HTC and Mask2Former, it can bring an improvement of 0.405\%. Finally, we achieve 63.008\% AP@0.50:0.95 on the test set of CVPR2022 AVA instance segmentation challenge.

\hfill

\section{Conclusion}

In this paper, we introduce the technical details of our submission to the CVPR2022 AVA Challenge, including the model and data augmentation strategy, the effective training strategy, and the integration of two different instance segmentation framework. Experimental results demonstrate that our approach can achieve a competitive result on the test set. Finally, our approach achieves 63.008\%AP@0.50:0.95 on the test set of CVPR2022 AVA Accessibility Vision and Autonomy Challenge.


{\small
\bibliographystyle{ieee_fullname}
\bibliography{egbib}

\begin{thebibliography}{10}\itemsep=-1pt

\bibitem{bodla2017soft}
Navaneeth Bodla, Bharat Singh, Rama Chellappa, and Larry~S Davis.
\newblock Soft-nms--improving object detection with one line of code.
\newblock In {\em Proceedings of the IEEE international conference on computer
  vision}, pages 5561--5569, 2017.

\bibitem{carion2020end}
Nicolas Carion, Francisco Massa, Gabriel Synnaeve, Nicolas Usunier, Alexander
  Kirillov, and Sergey Zagoruyko.
\newblock End-to-end object detection with transformers.
\newblock In {\em European conference on computer vision}, pages 213--229.
  Springer, 2020.

\bibitem{chen2019hybrid}
Kai Chen, Jiangmiao Pang, Jiaqi Wang, Yu Xiong, Xiaoxiao Li, Shuyang Sun,
  Wansen Feng, Ziwei Liu, Jianping Shi, Wanli Ouyang, et~al.
\newblock Hybrid task cascade for instance segmentation.
\newblock In {\em Proceedings of the IEEE/CVF Conference on Computer Vision and
  Pattern Recognition}, pages 4974--4983, 2019.

\bibitem{chen2019tensormask}
Xinlei Chen, Ross Girshick, Kaiming He, and Piotr Doll{\'a}r.
\newblock Tensormask: A foundation for dense object segmentation.
\newblock In {\em Proceedings of the IEEE/CVF International Conference on
  Computer Vision}, pages 2061--2069, 2019.

\bibitem{cheng2021mask2former}
Bowen Cheng, Ishan Misra, Alexander~G. Schwing, Alexander Kirillov, and Rohit
  Girdhar.
\newblock Masked-attention mask transformer for universal image segmentation.
\newblock 2022.

\bibitem{devlin2018bert}
Jacob Devlin, Ming-Wei Chang, Kenton Lee, and Kristina Toutanova.
\newblock Bert: Pre-training of deep bidirectional transformers for language
  understanding.
\newblock {\em arXiv preprint arXiv:1810.04805}, 2018.

\bibitem{9578639}
Golnaz Ghiasi, Yin Cui, Aravind Srinivas, Rui Qian, Tsung-Yi Lin, Ekin~D.
  Cubuk, Quoc~V. Le, and Barret Zoph.
\newblock Simple copy-paste is a strong data augmentation method for instance
  segmentation.
\newblock In {\em 2021 IEEE/CVF Conference on Computer Vision and Pattern
  Recognition (CVPR)}, pages 2917--2927, 2021.

\bibitem{he2017mask}
Kaiming He, Georgia Gkioxari, Piotr Doll{\'a}r, and Ross Girshick.
\newblock Mask r-cnn.
\newblock In {\em Proceedings of the IEEE international conference on computer
  vision}, pages 2961--2969, 2017.

\bibitem{Liang2021CBNetV2AC}
Tingting Liang, Xiao Chu, Yudong Liu, Yongtao Wang, Zhi Tang, Wei Chu, Jingdong
  Chen, and Haibin Ling.
\newblock Cbnetv2: A composite backbone network architecture for object
  detection.
\newblock {\em ArXiv}, abs/2107.00420, 2021.

\bibitem{liu2018path}
Shu Liu, Lu Qi, Haifang Qin, Jianping Shi, and Jiaya Jia.
\newblock Path aggregation network for instance segmentation.
\newblock In {\em Proceedings of the IEEE conference on computer vision and
  pattern recognition}, pages 8759--8768, 2018.

\bibitem{liu2021swin}
Ze Liu, Yutong Lin, Yue Cao, Han Hu, Yixuan Wei, Zheng Zhang, Stephen Lin, and
  Baining Guo.
\newblock Swin transformer: Hierarchical vision transformer using shifted
  windows.
\newblock In {\em Proceedings of the IEEE/CVF International Conference on
  Computer Vision}, pages 10012--10022, 2021.

\bibitem{loshchilov2018decoupled}
Ilya Loshchilov and Frank Hutter.
\newblock Decoupled weight decay regularization.
\newblock In {\em International Conference on Learning Representations}, 2018.

\bibitem{radford2019language}
Alec Radford, Jeffrey Wu, Rewon Child, David Luan, Dario Amodei, Ilya
  Sutskever, et~al.
\newblock Language models are unsupervised multitask learners.
\newblock {\em OpenAI blog}, 1(8):9, 2019.

\bibitem{ramachandran2019stand}
Prajit Ramachandran, Niki Parmar, Ashish Vaswani, Irwan Bello, Anselm Levskaya,
  and Jon Shlens.
\newblock Stand-alone self-attention in vision models.
\newblock {\em Advances in Neural Information Processing Systems}, 32, 2019.

\bibitem{wang2020solo}
Xinlong Wang, Tao Kong, Chunhua Shen, Yuning Jiang, and Lei Li.
\newblock Solo: Segmenting objects by locations.
\newblock In {\em European Conference on Computer Vision}, pages 649--665.
  Springer, 2020.

\bibitem{wang2020solov2}
Xinlong Wang, Rufeng Zhang, Tao Kong, Lei Li, and Chunhua Shen.
\newblock Solov2: Dynamic and fast instance segmentation.
\newblock {\em arXiv preprint arXiv:2003.10152}, 2020.

\bibitem{wang2020centermask}
Yuqing Wang, Zhaoliang Xu, Hao Shen, Baoshan Cheng, and Lirong Yang.
\newblock Centermask: single shot instance segmentation with point
  representation.
\newblock In {\em Proceedings of the IEEE/CVF conference on computer vision and
  pattern recognition}, pages 9313--9321, 2020.

\bibitem{zhang2020swa}
Haoyang Zhang, Ying Wang, Feras Dayoub, and Niko S{\"u}nderhauf.
\newblock Swa object detection.
\newblock {\em arXiv preprint arXiv:2012.12645}, 2020.

\bibitem{9710390}
J. Zhang, M. Zheng, M. Boyd, and E. Ohn-Bar.
\newblock X-world: Accessibility, vision, and autonomy meet.
\newblock In {\em 2021 IEEE/CVF International Conference on Computer Vision
  (ICCV)}, pages 9742--9751, Los Alamitos, CA, USA, oct 2021. IEEE Computer
  Society.

\bibitem{Zhao_2020_CVPR}
Hengshuang Zhao, Jiaya Jia, and Vladlen Koltun.
\newblock Exploring self-attention for image recognition.
\newblock In {\em Proceedings of the IEEE/CVF Conference on Computer Vision and
  Pattern Recognition (CVPR)}, June 2020.

\end{thebibliography}
}

\end{document}